\documentclass[conference]{IEEEtran}
\IEEEoverridecommandlockouts

\usepackage{cite}
\usepackage{amsmath,amssymb,amsfonts}
\usepackage{algorithmic}
\usepackage{graphicx}
\usepackage{textcomp}
\usepackage{xcolor}
\usepackage{multirow}
\usepackage{comment}
\usepackage{tabularx}
\usepackage{amsmath}
\usepackage{array}
\usepackage{float}
\usepackage[colorlinks = true, citecolor = black, urlcolor = gray, linkcolor = black]{hyperref}

\begin{document}

\title{Analysis of Argument Structure Constructions in the Large Language Model BERT}


\author{
Pegah Ramezani\textsuperscript{1,2}, 
Achim Schilling\textsuperscript{2,3},
Patrick Krauss\textsuperscript{2,3} \\
\textsuperscript{1}Department of English and American Studies, University Erlangen-Nuremberg, Germany\\
\textsuperscript{2}Cognitive Computational Neuroscience Group, Pattern Recognition Lab, University Erlangen-Nuremberg, Germany\\
\textsuperscript{3}Neuroscience Lab, University Hospital Erlangen, Germany\\
pegah.ramezani@fau.de, achim.schilling@fau.de, patrick.krauss@fau.de
}

\maketitle

\begin{abstract}
Understanding how language and linguistic constructions
are processed in the brain is a fundamental question
in cognitive computational neuroscience. In this study, we investigate the processing and representation of Argument Structure Constructions (ASCs) in the BERT language model, extending previous analyses conducted with Long Short-Term Memory (LSTM) networks. We utilized a custom GPT-4 generated dataset comprising 2000 sentences, evenly distributed among four ASC types: transitive, ditransitive, caused-motion, and resultative constructions. BERT was assessed using the various token embeddings across its 12 layers.
Our analyses involved visualizing the embeddings with Multidimensional Scaling (MDS) and t-Distributed Stochastic Neighbor Embedding (t-SNE), and calculating the Generalized Discrimination Value (GDV) to quantify the degree of clustering. We also trained feedforward classifiers (probes) to predict construction categories from these embeddings.
Results reveal that CLS token embeddings cluster best according to ASC types in layers 2, 3, and 4, with diminished clustering in intermediate layers and a slight increase in the final layers. Token embeddings for DET and SUBJ showed consistent intermediate-level clustering across layers, while VERB embeddings demonstrated a systematic increase in clustering from layer 1 to 12. OBJ embeddings exhibited minimal clustering initially, which increased substantially, peaking in layer 10.
Probe accuracies indicated that initial embeddings contained no specific construction information, as seen in low clustering and chance-level accuracies in layer 1. From layer 2 onward, probe accuracies surpassed 90 percent, highlighting latent construction category information not evident from GDV clustering alone.
Additionally, Fisher Discriminant Ratio (FDR) analysis of attention weights revealed that OBJ tokens had the highest FDR scores, indicating they play a crucial role in differentiating ASCs, followed by VERB and DET tokens. SUBJ, CLS, and SEP tokens did not show significant FDR scores.
Our study underscores the complex, layered processing of linguistic constructions in BERT, revealing both similarities and differences compared to recurrent models like LSTMs. Future research will compare these computational findings with neuroimaging data during continuous speech perception to better understand the neural correlates of ASC processing.
This research demonstrates the potential of both recurrent and transformer-based neural language models to mirror linguistic processing in the human brain, offering valuable insights into the computational and neural mechanisms underlying language understanding.
\end{abstract}



\section*{Introduction}

Understanding how the brain processes and represents language is a fundamental challenge in cognitive neuroscience \cite{pulvermuller2002neuroscience}. This paper adopts a usage-based constructionist approach, which views language as a system of form-meaning pairs (constructions) that link patterns to specific communicative functions \cite{goldberg2009nature, goldberg2003constructions}. Argument Structure Constructions (ASCs), such as transitive, ditransitive, caused-motion, and resultative constructions, are particularly important for language comprehension and production \cite{goldberg1995constructions, goldberg2006constructions, goldberg2019explain}. These constructions are key to syntactic theory and essential for constructing meaning in sentences. Exploring the neural and computational mechanisms underlying the processing of these constructions can yield significant insights into language and cognition \cite{pulvermuller2012meaning, pulvermuller2021biological, henningsen2022modelling, pulvermuller2023neurobiological}.

In recent years, advances in computational neuroscience have enabled the use of artificial neural networks to model various aspects of human cognition \cite{cohen2022recent}. Furthermore, the synergy between AI and cognitive neuroscience has led to a better understanding of the brain's unique complexities \cite{krauss2024artificial}. AI models, inspired by neural networks \cite{hassabis2017neuroscience}, have allowed neuroscientists to delve deeper into the brain's workings, offering insights that were previously unattainable \cite{krauss2023kunstliche}. These models have been particularly useful in studying how different parts of the brain interact and process information \cite{savage2019ai}. 

Among these neural network models, recurrent neural networks (RNNs) \cite{krauss2019weight, metzner2022dynamics, metzner2024quantifying}, and specifically Long Short-Term Memory (LSTM) networks \cite{hochreiter1997long}, have shown considerable promise in modeling sequential data, such as natural language \cite{wang2015learning}. However, transformer based large language models (LLM) like ChatGPT \cite{vaswani2017attention, radford2018improving} and BERT (Bidirectional Encoder Representations from Transformers) \cite{devlin2018bert} have shown remarkable capabilities in understanding and generating human language.

In previous studies using RNNs, particularly LSTM networks, we have demonstrated the emergence of representations for word classes and syntactic rules in the hidden layer activation of such networks when trained on next-word prediction tasks \cite{surendra2023word}. Furthermore, we showed that recurrent language models effectively differentiate between various Argument Structure Constructions (ASCs), forming distinct clusters for each ASC type in their internal representations, with the most pronounced clustering in the final hidden layer \cite{ramezani2024analysis}. These findings suggest that neural language models can capture complex linguistic patterns, making them valuable tools and models for studying language processing in the brain. While capturing lexico-semantic information is essential, interpreting the meanings of constructions can enhance the human-likeness of these models. Given that LLMs undergo extensive training on vast datasets, they are expected to effectively grasp human linguistic knowledge.

In this study, we extend our previous analyses of LSTM networks by investigating how ASCs are processed and represented in a large language model (LLM), in particular BERT, which, with its bidirectional attention mechanism, allows for a deeper and more nuanced understanding of linguistic context compared to traditional RNNs. By examining BERT's internal representations across its multiple layers, we aim to uncover how different ASCs are encoded and whether these representations align with those observed in LSTM networks.

To this end, we utilized a custom dataset generated by GPT-4, consisting of 2000 sentences evenly distributed among four ASC types: transitive, ditransitive, caused-motion, and resultative constructions. We analyzed the embeddings produced by BERT's CLS token and specific token embeddings (DET, SUBJ, VERB, OBJ) across its 12 layers. Our methodology involved visualizing these embeddings using Multidimensional Scaling (MDS) and t-Distributed Stochastic Neighbor Embedding (t-SNE), calculating the Generalized Discrimination Value (GDV) to quantify clustering, and employing feedforward classifiers (probes) to predict construction categories from the embeddings.

Our findings reveal distinct patterns of clustering and information encoding across BERT's layers, highlighting the model's ability to capture complex linguistic constructions. 

These results are compared to those from LSTM-based models, providing a comprehensive understanding of how different neural architectures process linguistic information. Future research will focus on validating these findings with larger language models and correlating them with neuroimaging data obtained during continuous speech perception, aiming to bridge the gap between computational models and neural mechanisms of language understanding.


\section*{Methods}

\subsection*{Dataset creation using GPT4}

To investigate the processing and representation of different Argument Structure Constructions (ASCs) in a recurrent neural language model, we created a custom dataset using GPT-4. This dataset was designed to include sentences that exemplify four distinct ASCs: transitive, ditransitive, caused-motion, and resultative constructions (cf. Table \ref{tab:agree_that_construction}). Each ASC category consisted of 500 sentences, resulting in a total of 2000 sentences.

\subsubsection*{Selection of Argument Structure Constructions}

The four ASCs selected for this study are foundational to syntactic theory and represent different types of sentence structures: \\
Transitive Constructions: Sentences where a subject performs an action on a direct object (e.g., "The cat chased the mouse"). \\
Ditransitive Constructions: Sentences where a subject performs an action involving a direct object and an indirect object (e.g., "She gave him a book"). \\
Caused-motion Constructions: Sentences where a subject causes an object to move in a particular manner (e.g., "He pushed the cart into the garage"). \\
Resultative Constructions: Sentences where an action results in a change of state of the object (e.g., "She painted the wall red").

\begin{table}[h]
\centering
\begin{tabular}{|p{0.20\linewidth}|p{0.28\linewidth}|p{0.28\linewidth}|}
\hline
\textbf{Constructions} & \textbf{Structure} & \textbf{Example} \\
\hline
Transitive & Subject + Verb + Object & The baker baked a cake. \\
\hline
Ditransitive & Subject + Verb + Object1 + Object2 & The teacher gave students homework. \\
\hline
Caused-Motion & Subject + Verb + Object + Path & The cat chased the mouse into the garden. \\
\hline
Resultative & Subject + Verb + Object + State & The chef cut the cake into slices. \\
\hline
\end{tabular}
\caption{Name, structure, and example of each construction}
\label{tab:agree_that_construction}
\end{table}

\subsubsection*{Generation of Sentences}

To ensure the diversity and quality of the sentences in our dataset, we utilized GPT-4, a state-of-the-art language model developed by OpenAI \cite{}. The generation process involved the following steps:
Prompt Design: We created specific prompts for GPT-4 to generate sentences for each ASC category. These prompts included example sentences and detailed descriptions of the desired sentence structures to guide the model in generating appropriate constructions.
Sentence Generation: Using the designed prompts, we generated 500 sentences for each ASC category. The generation process was carefully monitored to ensure that the sentences adhered to the syntactic patterns of their respective constructions.
Manual Review and Filtering: After the initial generation, we manually reviewed the sentences to ensure their grammatical correctness and adherence to the intended ASC types. Sentences that did not meet these criteria were discarded and replaced with newly generated ones.
Balancing the Dataset: To prevent any bias in the model training, we ensured that the dataset was balanced, with an equal number of sentences (500) for each of the four ASC categories.

\subsubsection*{Text Tokenization}

After tokenization using BERT's tokenizer, we ensured that the tokens of all sentences within each construction were identical. This standardization facilitated easier tracking and better comparison by focusing on differences across constructions rather than within them. The tokens used in our dataset include Subject (Subj), Verb (Verb), Direct Object (Obj), Indirect Object (IndObj), Object of Preposition (ObjPrep), Preposition (Prep), and Determiner (Det). Additionally, the CLS tokens were added by the BERT tokenizer for sentence classification and separation.

\begin{table}[h]
\centering
\begin{tabular}{|p{0.21\linewidth}|p{0.72\linewidth}|}
\hline
\textbf{Constructions} & \textbf{tokens} \\
\hline
Transitive & CLS +Det +Subj +Verb +Det +Obj +SEP \\
\hline
Ditransitive & CLS +Det +Subj +Verb +IndObj +Obj +SEP \\
\hline
Caused-Motion & CLS +Det +Subj +Verb +Det +Obj +Prep +Det +ObjPrep +SEP \\
\hline
Resultative & CLS +Det +Subj +Verb +Det +Obj +Prep +ObjPrep +SEP \\
\hline
Common & CLS +Det +Subj +Verb +Obj +SEP \\
\hline
\end{tabular}
\caption{Name and token of each construction}
\label{tab:agree_that_construction}
\end{table}

The resulting dataset, comprising 2000 sentences represented as token sequences, serves as a robust foundation for probing and analyzing the BERT model. This carefully curated and preprocessed dataset enables us to investigate how different ASCs are processed and represented within the BERT, providing insights into the underlying computational mechanisms.

For a subset of our analysis, we focused on common tokens across all constructions to enable a consistent comparison of single tokens within different ASCs. This approach ensured that our analysis captured the essential structural and functional aspects of each construction type, thereby providing a robust framework for understanding how BERT processes and represents linguistic constructions.

\subsection*{BERT architecture}

For our study, we utilized the BERT (Bidirectional Encoder Representations from Transformers) model, renowned for its ability to process bidirectional context effectively \cite{devlin2018bert}. BERT's architecture comprises multiple layers of bidirectional transformer encoders, which enable it to consider both left and right context at all layers, enhancing its performance on a range of natural language understanding tasks.

The BERT model starts with tokenization, where text is split into subword units using WordPiece tokenization, allowing the model to handle a diverse array of words and word forms efficiently. Special tokens CLS and SEP are added to the beginning and end of each input sequence, respectively. The CLS token is used for classification tasks and summarized the entire input, while the SEP token denotes sentence boundaries.

In the embedding layer, input tokens are converted into embeddings that combine token embeddings, segment embeddings, and position embeddings. These embeddings are then passed through multiple layers of transformer encoders. BERT's architecture includes 12 layers (in the base model) of transformer encoders, each comprising self-attention mechanisms and feedforward neural networks. Each encoder layer has multiple attention heads, allowing the model to focus on different parts of the input sequence simultaneously. The self-attention mechanism computes a representation of each token by considering the entire input sequence, capturing complex dependencies and relationships.

The output of each transformer encoder layer provides contextualized representations of the input tokens. For each token, the final layer's output represents its contextualized embedding, which incorporates information from the entire input sequence. The CLS token's final layer embedding is typically used for classification tasks, as it contains an aggregated representation of the entire sequence.

BERT was pre-trained on a large corpus using masked language modeling and next sentence prediction tasks, enabling it to learn a rich representation of language. For our specific task, we utilized the pre-trained BERT model and fine-tuned it on our custom dataset to capture the nuances of Argument Structure Constructions (ASCs).

By leveraging BERT's robust architecture, we aimed to gain insights into how different ASCs are represented and processed across its layers. This detailed examination of BERT's internal representations provided a comprehensive understanding of the model's ability to encode complex linguistic constructions, facilitating comparison with recurrent models like LSTMs and enhancing our knowledge of computational language processing.

\subsection*{Analysis of Hidden Layer Activation}

We assessed BERT's ability to differentiate between the various constructions by analyzing the activations of its hidden layers and attention weights. Initially, the dataset underwent processing through the "bert-base-uncased" model without any fine-tuning. The model comprises 12 hidden layers, each containing 768 neurons. For each token, the activity of each layer was extracted for further analysis.

Given the high dimensionality of these activations, direct visual inspection is not feasible. To address this, we employed dimensionality reduction techniques to project the high-dimensional activations into a two-dimensional space. By combining different visualization and quantitative techniques, we were able to assess the BERT's internal representations and its ability to differentiate between the various linguistic constructions.

\subsubsection*{Multidimensional Scaling (MDS)}

This technique was used to reduce the dimensionality of the hidden layer activations, preserving the pairwise distances between points as much as possible in the lower-dimensional space. In particular, MDS is an efficient embedding technique to visualize high-dimensional point clouds by projecting them onto a 2-dimensional plane. Furthermore, MDS has the decisive advantage that it is parameter-free and all mutual distances of the points are preserved, thereby conserving both the global and local structure of the underlying data \cite{torgerson1952multidimensional, kruskal1964nonmetric,kruskal1978multidimensional,cox2008multidimensional, metzner2021sleep, metzner2023extracting, metzner2022classification}. 

When interpreting patterns as points in high-dimensional space and dissimilarities between patterns as distances between corresponding points, MDS is an elegant method to visualize high-dimensional data. By color-coding each projected data point of a data set according to its label, the representation of the data can be visualized as a set of point clusters. For instance, MDS has already been applied to visualize for instance word class distributions of different linguistic corpora \cite{schilling2021analysis}, hidden layer representations (embeddings) of artificial neural networks \cite{schilling2021quantifying,krauss2021analysis}, structure and dynamics of highly recurrent neural networks \cite{krauss2019analysis, krauss2019recurrence, krauss2019weight, metzner2023quantifying}, or brain activity patterns assessed during e.g. pure tone or speech perception \cite{krauss2018statistical,schilling2021analysis}, or even during sleep \cite{krauss2018analysis,traxdorf2019microstructure,metzner2022classification,metzner2023extracting}. 
In all these cases the apparent compactness and mutual overlap of the point clusters permits a qualitative assessment of how well the different classes separate.

\subsubsection*{t-Distributed Stochastic Neighbor Embedding (t-SNE)}

This method further helped in visualizing the complex structures within the activations by emphasizing local similarities, allowing us to see the formation of clusters corresponding to different Argument Structure Constructions (ASCs). t-SNE is a frequently used method to generate low-dimensional embeddings of high-dimensional data \cite{van2008visualizing}. However, in t-SNE the resulting low-dimensional projections can be highly dependent on the detailed parameter settings \cite{wattenberg2016use}, sensitive to noise, and may not preserve, but rather often scramble the global structure in data \cite{vallejos2019exploring, moon2019visualizing}. Here, we set the perplexity (number of next neighbours taken into account) to 100.

\subsection*{Generalized Discrimination Value (GDV)}

To quantify the degree of clustering, we used the GDV as published and explained in detail in \cite{schilling2021quantifying}. This GDV provides an objective measure of how well the hidden layer activations cluster according to the ASC types, offering insights into the model's internal representations. Briefly, we consider $N$ points $\mathbf{x_{n=1..N}}=(x_{n,1},\cdots,x_{n,D})$, distributed within $D$-dimensional space. A label $l_n$ assigns each point to one of $L$ distinct classes $C_{l=1..L}$. In order to become invariant against scaling and translation, each dimension is separately z-scored and, for later convenience, multiplied with $\frac{1}{2}$:
\begin{align}
s_{n,d}=\frac{1}{2}\cdot\frac{x_{n,d}-\mu_d}{\sigma_d}.
\end{align}
Here, $\mu_d=\frac{1}{N}\sum_{n=1}^{N}x_{n,d}\;$ denotes the mean,\\ \\
and $\sigma_d=\sqrt{\frac{1}{N}\sum_{n=1}^{N}(x_{n,d}-\mu_d)^2}$ the standard deviation of dimension $d$. \\ \\
Based on the re-scaled data points $\mathbf{s_n}=(s_{n,1},\cdots,s_{n,D})$, we calculate the {\em mean intra-class distances} for each class $C_l$ 
\begin{align}
\bar{d}(C_l)=\frac{2}{N_l (N_l\!-\!1)}\sum_{i=1}^{N_l-1}\sum_{j=i+1}^{N_l}{d(\textbf{s}_{i}^{(l)},\textbf{s}_{j}^{(l)})},
\end{align}
and the {\em mean inter-class distances} for each pair of classes $C_l$ and $C_m$
\begin{align}
\bar{d}(C_l,C_m)=\frac{1}{N_l  N_m}\sum_{i=1}^{N_l}\sum_{j=1}^{N_m}{d(\textbf{s}_{i}^{(l)},\textbf{s}_{j}^{(m)})}.
\end{align}
Here, $N_k$ is the number of points in class $k$, and $\textbf{s}_{i}^{(k)}$ is the $i^{th}$ point of class $k$.
The quantity $d(\textbf{a},\textbf{b})$ is the euclidean distance between $\textbf{a}$ and $\textbf{b}$. Finally, the Generalized Discrimination Value (GDV) is calculated from the mean intra-class and inter-class distances  as follows:
\begin{align}
\mbox{GDV}=\frac{1}{\sqrt{D}}\left[\frac{1}{L}\sum_{l=1}^L{\bar{d}(C_l)}\;-\;\frac{2}{L(L\!-\!1)}\sum_{l=1}^{L-1}\sum_{m=l+1}^{L}\bar{d}(C_l,C_m)\right]
 \label{GDVEq}
\end{align}

\noindent whereas the factor $\frac{1}{\sqrt{D}}$ is introduced for dimensionality invariance of the GDV with $D$ as the number of dimensions.

\vspace{0.2cm}\noindent Note that the GDV is invariant with respect to a global scaling or shifting of the data (due to the z-scoring), and also invariant with respect to a permutation of the components in the $N$-dimensional data vectors (because the euclidean distance measure has this symmetry). The GDV is zero for completely overlapping, non-separated clusters, and it becomes more negative as the separation increases. A GDV of -1 signifies already a very strong separation.

\subsection*{Probes}

Probes, a technique from the mechanistic explainability area of AI, are utilized to analyze deep neural networks \cite{alain2018understanding}. They are commonly applied in the field of natural language processing \cite{belinkov-2022-probing}. Probes are typically small, neural network-based classifiers, usually implemented as shallow fully connected networks. They are trained on the activations of specific neurons or layers of a larger neural network to predict certain features, which are generally believed to be necessary or beneficial for the network's task. If probes achieve accuracy higher than chance, it suggests that the information about the feature, or something correlated to it, is present in the activations.

Here, we employed edge probing to analyze different tokens using the methodology described by Tenney et al. \cite{tenney2019you}. This probing approach involves designing a classification model tailored to classify the hidden layer activities based on constructions. The model is systematically trained on a per-layer and per-token basis, targeting specific linguistic elements such as the CLS token, subject, and verb. This allows for detailed insights into how BERT encodes different Argument Structure Constructions (ASCs) across its layers.

The classification model used in this probing endeavor is a 4-class Support Vector Machine (SVM) classifier with a linear kernel. The SVM takes the hidden layer activity of a layer per token and predicts the class of its construction. This straightforward yet effective approach enables us to quantify the degree of clustering and construction-specific information present in different layers of BERT.

By training the SVM classifier on the hidden layer activations for various tokens, we can evaluate the model's performance in distinguishing between the four ASC types. In particular, an accuracy significantly above chance level indicates that information about the construction category is represented (latent) in the respective token embedding. The results from this probing technique provide a quantitative measure of classification performance and clustering tendencies, offering a comprehensive understanding of how linguistic constructions are represented within the BERT model.

\subsection*{Analysis of attention heads }
In BERT, each of the 12 layers contains 12 attention heads. For each head, there are attention weights for all tokens in the sequence relative to every other token. To facilitate a comparable analysis, we focused on the attention weights for the common tokens: CLS, DET, SUBJ, VERB, and OBJ.

This analysis aimed to identify which attention heads and layers exhibit the most significant differences among the four Argument Structure Constructions (ASCs). We then examined these attention heads in detail, evaluating their function and the weights assigned to each token.

To determine which tokens had more distinct weights across the constructions, we first summed all attention weights directed at each token from all other tokens. Next, we considered the attention weight of each token per head and layer as a feature. We then calculated the F-statistic using ANOVA (Analysis of Variance) to assess the variability of attention weights among the four constructions. A higher F-score indicates a greater difference in attention weights among the constructions.

Finally, we averaged the attention weights for each token across the heads and layers to provide a comprehensive view of the attention distribution. This multi-step approach allowed us to identify key attention heads and layers that significantly contribute to differentiating the ASCs, offering insights into the role of attention mechanisms in BERT's processing of linguistic constructions.

\subsection*{Fisher Discriminant Ratio (FDR)} 

The Fisher Discriminant Ratio (FDR) is a measure used in pattern recognition, feature selection, and machine learning to evaluate the discriminatory power of a feature \cite{kim2005robust, wang2011feature}. It helps determine how well a feature can distinguish between different classes. The FDR is calculated as the ratio of the variance between classes to the variance within classes. A higher FDR indicates that the feature has a greater ability to differentiate between classes.

In this study, we utilized the FDR to assess the attention weights in BERT for distinguishing between different Argument Structure Constructions (ASCs). By calculating the FDR for attention weights across each layer, we aimed to identify which layers and heads provide the most distinct representations of the ASCs.

The FDR was computed using the following formula:

\[
\text{FDR} = \frac{(\mu_1 - \mu_2)^2}{\sigma_1^2 + \sigma_2^2}
\]

where:
\begin{itemize}
    \item $\mu_1$ and $\mu_2$ are the means of the feature for class 1 and class 2, respectively.
    \item $\sigma_1^2$ and $\sigma_2^2$ are the variances of the feature for class 1 and class 2, respectively.
\end{itemize}

\subsection*{Code implementation, Computational resources, and programming libraries}

\subsection*{Code implementation, Computational resources, and programming libraries}

All simulations were run on a standard personal computer. The evaluation software was based on Python 3.9.13 \cite{oliphant2007python}. For matrix operations the numpy-library \cite{van2011numpy} was used and data visualization was done using matplotlib \cite{hunter2007matplotlib} and the seaborn library \cite{waskom2021seaborn}. The dimensionality reduction through MDS and t-SNE was done using the sci-kit learn library. Mathematical operations were performed with numpy \cite{numpy} and scikit-learn \cite{scikit-learn} libraries.
Visualizations were realized with matplotlib \cite{matplot} and networkX \cite{networkX}. For natural language processing we used SpaCy \cite{explosion2017spacy}.


\section*{Results}

To understand how the BERT model differentiates between various Argument Structure Constructions (ASCs), we visualized the activations of its hidden layers using Multidimensional Scaling (MDS) and t-Distributed Stochastic Neighbor Embedding (t-SNE). Additionally, we quantified the degree of clustering using the Generalized Discrimination Value (GDV).
Furthermore, we utilized probes to test for latent representations in the token embeddings, Finally, we assessed the attention heads and their discriminative power according to ASCs.

\subsection*{Hidden Layer activity cluster analysis}

\begin{figure}[t]
  \centering
  \includegraphics[width=\columnwidth]{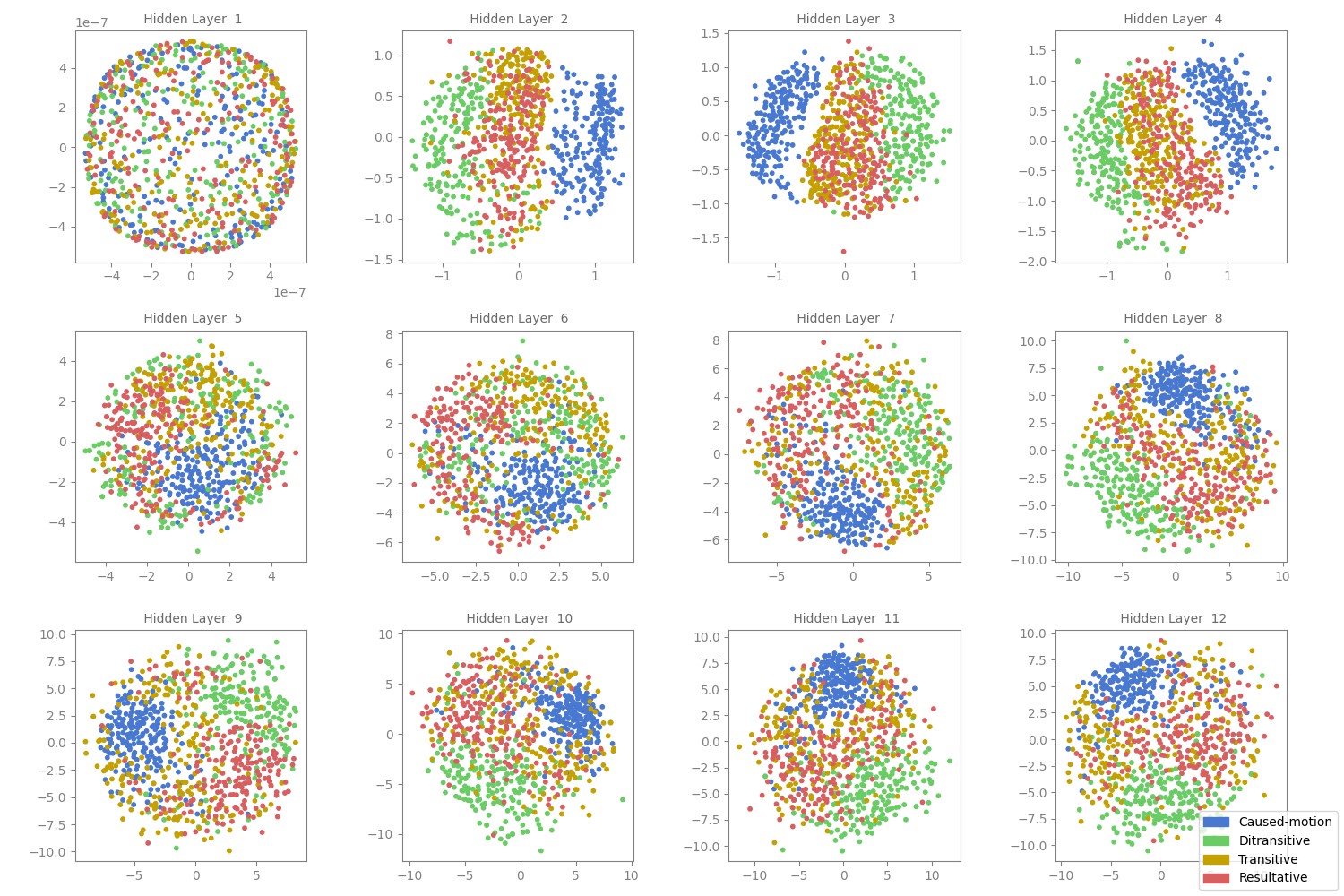}
  \caption{MDS projections of the CLS token embedding, i.e. hidden layer activation, from all hidden layers of the BERT model. Each point represents the activation of a sentence, color-coded according to its ASC type: caused-motion (blue), ditransitive (green), transitive (orange), and resultative (red).}
  \label{fig:MDS}
\end{figure}

Figure \ref{fig:MDS} shows the MDS projections of the CLS token embeddings from various layers of the BERT model. Each point represents the embedding of a sentence's CLS token. In the initial layer, there is minimal separation between the different ASC types, indicating that the input embeddings do not yet contain specific information about the construction categories.

As we move to the second layer, the separation between ASC types becomes more apparent, with distinct clusters forming for each construction type. This trend continues in the third and fourth layers, where the clustering is most pronounced. The inter-cluster distances increase, showing clearer differentiation between the ASC types. However, in these middle layers, there is still some overlap, particularly between the ditransitive and resultative constructions.

In layers five, six, and seven, the degree of clustering decreases slightly, with the clusters becoming less distinct. This reduction in clustering suggests a transformation in how BERT processes and integrates contextual information across these layers.

Interestingly, in the later layers (eight to twelve), there is a slight increase in the degree of clustering again. The clusters for the different ASC types become more defined compared to the intermediate layers, indicating a resurgence in the model's ability to distinguish between the construction types. This pattern suggests that BERT refines its understanding and representation of linguistic constructions in the deeper layers.

Overall, the CLS token embeddings demonstrate varying degrees of clustering across the BERT layers, with the best separation observed in the early layers (2-4) and a notable refinement in the final layers (8-12). This analysis reveals the complex and layered nature of how BERT processes linguistic constructions, highlighting the model's capability to encode and differentiate between ASCs at multiple stages of its architecture.

\begin{figure}[H]
  \centering
  \includegraphics[width=\columnwidth]{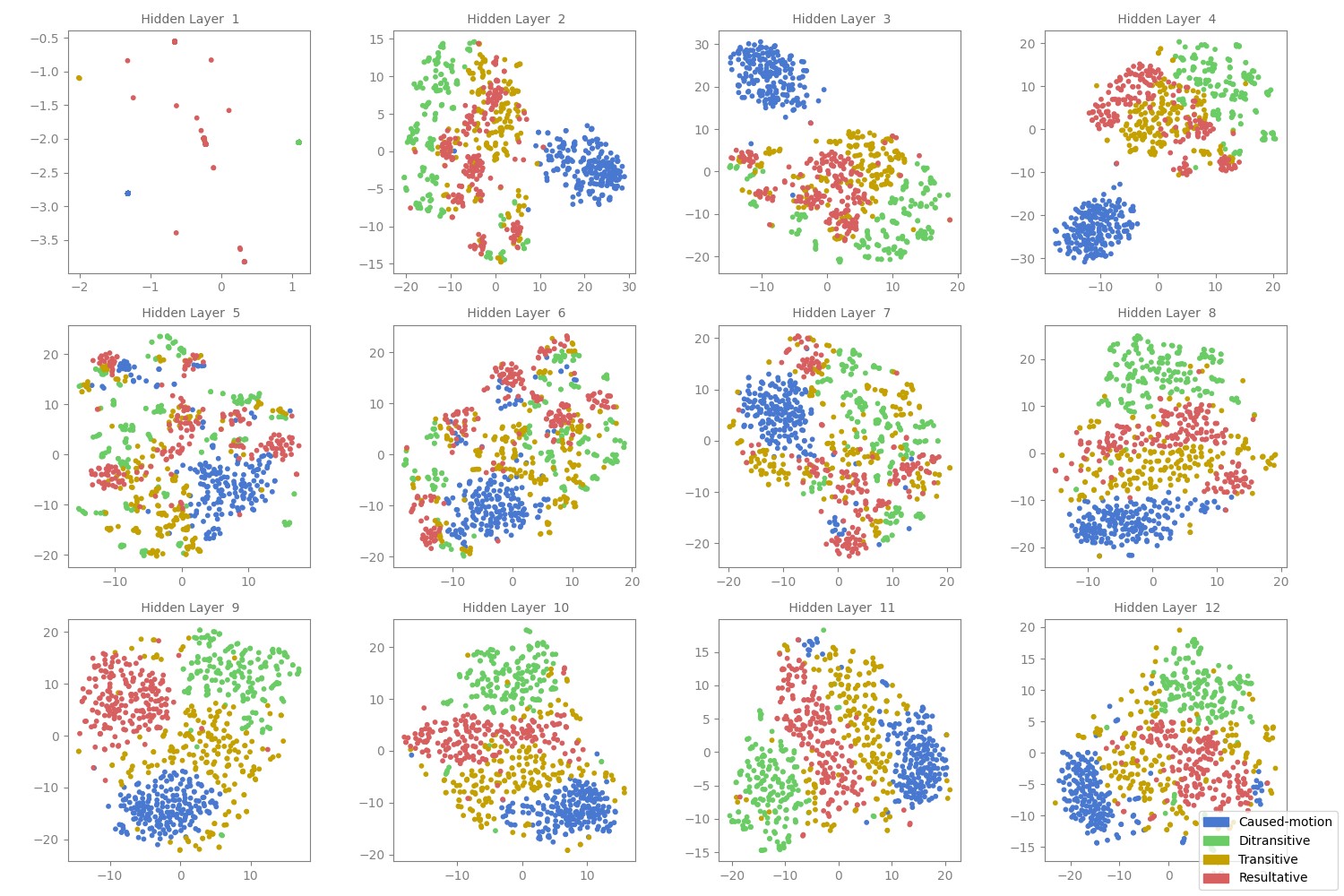}
  \caption{t-SNE projections of the CLS token embedding from all hidden layers of the BERT model. Each point represents the activation of a sentence, color-coded according to its ASC type: caused-motion (blue), ditransitive (green), transitive (orange), and resultative (red).}
  \label{fig:tSNE}
\end{figure}

The corresponding t-SNE projections shown in Figure \ref{fig:tSNE} show results similar to the MDS projections but with more detailed sub-cluster structures. Again, each point in the t-SNE plot represents the embedding of a sentence's CLS token. In the initial layer, minimal separation between ASC types is observed, aligning with the MDS results. Layers two, three, and four show distinct clusters, while layers five to seven exhibit reduced cluster definition. In the later layers (eight to twelve), clearer clustering re-emerges. Although, the t-SNE plots reveal nuanced sub-structures within clusters, it remains uncertain whether these sub-cluster structures are real effects or artifacts of t-SNE.

\begin{figure}[H]
  \centering
  \includegraphics[width=\columnwidth]{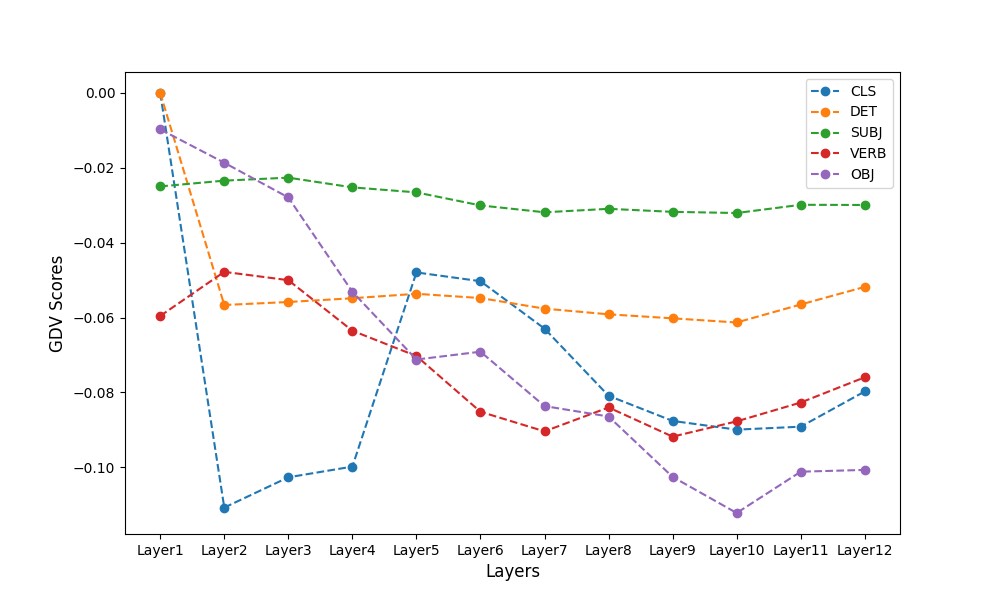}
  \caption{GDV score of hidden layer activations. Note that, lower GDV values indicate better-defined clusters. The qualitative results from the MDS and t-SNE projections of the CLS token embeddings are underpinned by the GDV with best clustering occurring in layer 2. The GDV of specific token embeddings reveals distinct patterns of clustering across the BERT layers. DET and SUBJ token embeddings exhibited relatively constant clustering at an intermediate level across all layers, capturing construction-specific information consistently. VERB token embeddings showed a slight but systematic increase in clustering from layer 1 to layer 12, indicating improved differentiation according to construction types in deeper layers. OBJ token embeddings began with no clustering in layer 1 but significantly increased across layers, reaching a clustering level in layer 10 comparable to the CLS token in layer 2. These results highlight the varying contributions of different tokens to the representation of ASCs within BERT, with some tokens showing dynamic changes and others maintaining consistent clustering.}
  \label{fig:GDV}
\end{figure}

To quantitatively assess the clustering quality, we calculated the GDV for the CLS token activations of each hidden layer (cf. Figure \ref{fig:GDV}). Lower GDV values indicate better defined clusters. The qualitative results of the MDS and t-SNE projections of the CLS token embeddings are supported by the GDV. However, the GDV of specific token embeddings reveals distinct patterns of clustering across the BERT layers.

The embeddings for DET and SUBJ tokens exhibited relatively constant clustering across all layers, maintaining an intermediate level of separation between the different Argument Structure Constructions (ASCs). This consistent clustering indicates that these tokens capture construction-specific information throughout the layers of the model.

The VERB token embeddings showed a slight but systematic increase in clustering from layer 1 to layer 12. Starting at an intermediate level, the clustering gradually improved across layers, suggesting that BERT increasingly differentiates the VERB token embeddings according to construction types as the model processes deeper layers.

The OBJ token embeddings began at a very low clustering level in layer 1, indicating no initial differentiation among the construction types. However, as the layers progressed, the clustering of OBJ token embeddings significantly increased. By layer 10, the degree of clustering for OBJ tokens reached a level comparable to that of the CLS token in layer 2, demonstrating a marked improvement in distinguishing the construction categories.

These GDV results highlight how different tokens contribute to the representation of ASCs within BERT. The findings suggest that while some tokens like DET and SUBJ consistently capture construction-specific information, others like VERB and OBJ show more dynamic changes in clustering, reflecting the layered and evolving nature of BERT's processing.

\subsection*{Hidden Layer activity Probing}

\begin{figure}[H]
  \centering
  \includegraphics[width=\columnwidth]{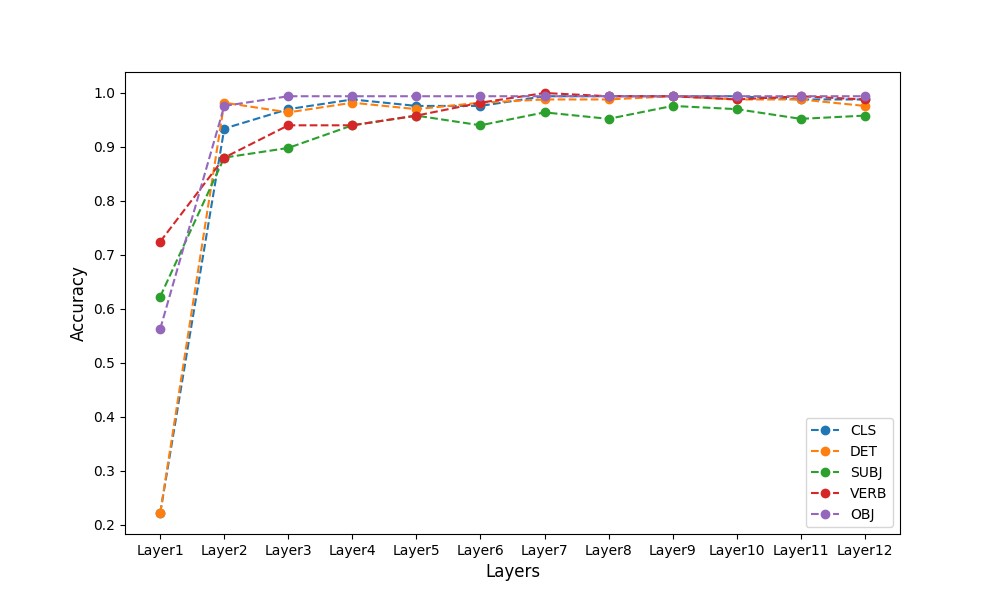}
  \caption{Accuracies of probing of hidden layers for classification of constructions per common tokens. Probe accuracies for CLS, DET, SUBJ, VERB, and OBJ tokens start at low or chance levels in layer 1, indicating that the initial embeddings contain no specific information about construction type, as also revealed by the GDV cluster analysis. However, from layer 2 to layer 12, all probe accuracies for different tokens consistently exceeded 90 percent, indicating latent information about construction categories in all token embeddings, even when not revealed through clustering alone.}
  \label{fig:probes}
\end{figure}

The probing analysis involved training a 4-class Support Vector Machine (SVM) classifier with a linear kernel to classify hidden layer activities based on Argument Structure Constructions (ASCs). This classifier was systematically trained on a per-layer and per-token basis, targeting specific linguistic elements such as the CLS token, subject (SUBJ), verb (VERB), and object (OBJ). The results are summarized in Figure \ref{fig:probes}.

In the initial layer, the probe accuracy for the CLS token was at chance level (25 percent), indicating that the input embeddings did not contain specific information about construction categories. From layer 2 onwards, the probe accuracy for the CLS token consistently exceeded 90 percent, demonstrating that construction-specific information becomes latent in the CLS token embeddings early in the processing. Probe accuracy slightly decreased in intermediate layers (5 to 7) but increased again in the later layers (8 to 12), showing a resurgence of construction-specific information.

Probe accuracies for DET and SUBJ tokens also started at chance levels in layer 1, indicating no specific information about construction categories. However, from layer 2 onwards, the accuracies consistently exceeded 90 precent, suggesting that these tokens capture construction-specific information effectively throughout the model's layers.

The probe accuracy for VERB tokens started at a low level in layer 1 but showed a systematic increase, with accuracies surpassing 90 percent from layer 2 to layer 12. This indicates that BERT progressively improves its differentiation of VERB token embeddings according to construction types in deeper layers.

Probe accuracy for OBJ tokens began at a very low level in layer 1, reflecting no initial differentiation among the construction types. However, as layers progressed, the probe accuracy for OBJ tokens significantly increased, reaching and maintaining levels above 90 precent from layer 2 to layer 12, demonstrating a marked improvement in distinguishing construction categories for OBJ tokens.

These probing results reveal that probe accuracies for CLS, DET, SUBJ, VERB, and OBJ tokens start at low or chance levels in layer 1, indicating that the initial embeddings contain no specific information about construction type, as also revealed by the GDV cluster analysis. However, from layer 2 to layer 12, all probe accuracies for different tokens consistently exceeded 90 percent indicating latent information about construction categories in all token embeddings, even when not revealed through clustering alone.

\subsection*{Attention weight analysis}

\begin{figure}[H]
  \centering
  \includegraphics[width=\columnwidth]{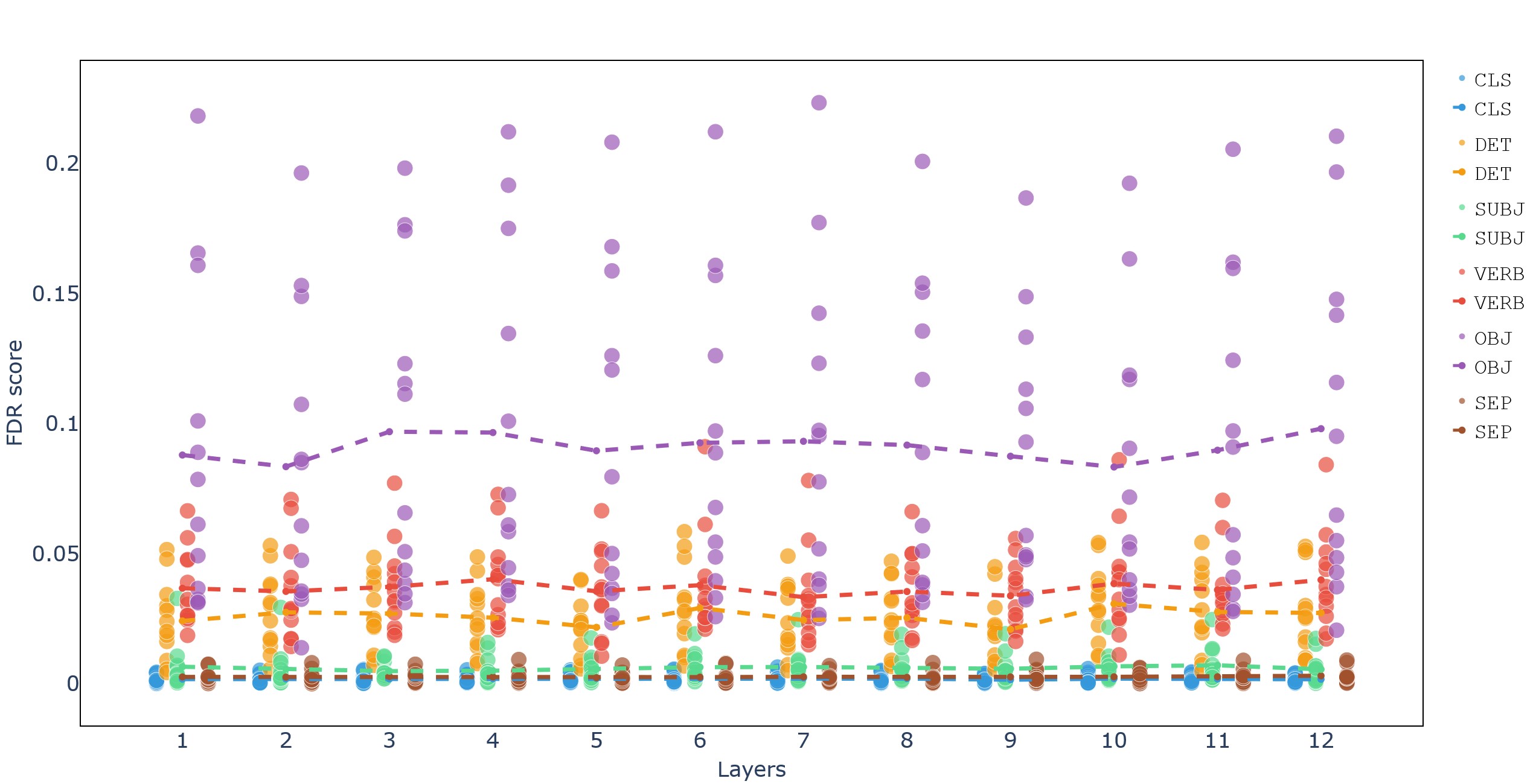}
  \caption{Fisher Discriminant Ratio (FDR) scores for each token across all layers and attention heads. Each dot represents the FDR score of a specific attention head, while the dashed line indicates the mean FDR of all attention heads. Layers with similar FDR scores suggest consistent patterns of attention across those layers.}
  \label{fig:attention}
\end{figure}

In Figure \ref{fig:attention} the Fisher Discriminant Ratio (FDR) scores for each token across all layers and attention heads are shown. The analysis reveals that the OBJ token has the highest FDR scores across all layers, indicating that this token plays a crucial role in differentiating the four Argument Structure Constructions (ASCs). The prominence of the OBJ token suggests it is key to distinguishing between the construction types.

The VERB token is the second most significant, showing high FDR scores in the same heads where the OBJ token performs well. This indicates that the verb token also contributes substantially to the differentiation of the constructions.

The DET token follows in significance. Despite its form being similar across all constructions, its embedding captures contextual information that aids in distinguishing the construction types.

In contrast, the SUBJ and CLS tokens exhibit no notable FDR scores, indicating that these tokens do not significantly contribute to the differentiation of the constructions.

This attention weight analysis highlights the critical role of the OBJ and VERB tokens in distinguishing between different ASCs within BERT's attention mechanisms, with the DET token also playing a meaningful, albeit lesser, role.


\section*{Discussion}

\subsection*{Summary of Findings}

In this study, we investigated how different Argument Structure Constructions (ASCs) are processed and represented within the BERT language model. Utilizing a custom GPT-4 generated dataset consisting of sentences across four ASC types (transitive, ditransitive, caused-motion, and resultative constructions), we analyzed BERT's internal representations and attention mechanisms using various techniques, including MDS, t-SNE, Generalized Discrimination Value (GDV), probing, and Fisher Discriminant Ratio (FDR) analysis.

Our results revealed distinct patterns in how BERT processes ASCs. Specifically, the CLS token embeddings exhibited clear clustering in layers 2, 3, and 4, with clustering quality decreasing in intermediate layers and improving again in later layers. This suggests a complex, layered approach to representing ASCs within BERT. The specific token analysis showed that DET and SUBJ tokens maintained intermediate-level clustering consistently across layers, while VERB and OBJ tokens displayed more dynamic changes, with OBJ tokens showing a marked improvement in clustering in deeper layers.

Among the four constructions we examined, distinguishing between transitive and resultative constructions proved to be more challenging for BERT. This similarity is evident in two primary ways. First, the visualization of dimensionally reduced hidden layer activity, particularly in MDS, shows significant overlap between the data points for transitive and resultative constructions. Second, the confusion matrix for the classification of the CLS token reveals that most errors involve misclassifying these two constructions as each other. This can be explained by noting that, in our dataset, resultative sentences without their final state resemble transitive sentences. For instance, "The artist painted the wall blue" (resultative) becomes "The artist painted the wall" (transitive) when the final state is removed.

Probing results indicated that probe accuracies for all tokens were at chance levels in layer 1, but from layer 2 to layer 12, all tokens achieved accuracies above 90 percent. This indicates that latent information about construction categories is embedded in token representations early on and remains robust throughout the model's layers.

At the second layer, the performance of tokens becomes more similar. This occurs because the embeddings are influenced not only by the tokens themselves but also by the general understanding of the sentences. Consequently, the performance of all tokens improves, and interestingly, the accuracy of the CLS and DET tokens, which was initially quite low, begins to increase.

Our analysis of token accuracy across layers revealed that the first layer primarily decodes lexical information, resulting in low accuracy for context-dependent tokens like CLS and DET. However, as we move to higher layers, token performance improves, reflecting BERT's increasing ability to leverage general sentence understanding. This improvement underscores that distinguishing constructions relies not only on lexical and syntactic information but also on the broader semantic context.

In summary, we believe that the high accuracy and low GDV (since they are indirectly related) in the first layer indicate how specific each token is to a construction. The results show that the verb is most specific, followed by the subject and object tokens, which are specific to certain aspects as well.

The FDR analysis of attention weights highlighted that the OBJ token had the highest FDR scores, suggesting it is key to differentiating the four ASCs. The VERB token also showed significant FDR scores, followed by the DET token, which, despite its consistent form, captured contextually relevant information. In contrast, the SUBJ and CLS tokens did not contribute significantly to the differentiation of constructions.

The result of FDR analysis for attention heads shows that different layers have slightly similar functions regarding attention heads. Notably, the sum of weights for the Object token differs the most among all constructions. This finding contrasts with the results of hidden layer activity, where the Verb token was the most distinct. The second most distinct token in the FDR analysis is the Verb, followed by DET, which maintains the same score even in the first layer. After these, the CLS, and SUBJ tokens have lower scores.

Furthermore, the FDR analysis of attention heads showed that different layers have similar functions, with the Object token displaying the most variability across constructions. This contrasts with hidden layer activity, where the Verb token was most distinct. The alignment of attention activity and hidden layer activity, despite their independent functions, highlights BERT's robust performance in understanding constructions.

\subsection*{Implications and Comparisons with Previous Studies}

Our findings align with previous studies on recurrent neural networks (RNNs) like LSTMs, which demonstrated that simple, brain-constrained models could effectively distinguish between different linguistic constructions. However, BERT's transformer-based architecture provides a more nuanced and multi-layered representation of ASCs, as evidenced by the dynamic changes in clustering and probing accuracies across layers.

The role of the verb token in constructions has been discussed in several studies. Some studies argue that verbs are construction-specific; for example, the verb 'visit' is lexically specified as being transitive \cite{fujita1989knowledge}. Conversely, construction grammar suggests that constructions do not depend on specific verbs \cite{goldberg1995constructions}. For instance, the verb "cut" can be used in both transitive constructions like "Bob cut the bread" and ditransitive constructions like "Bob cut Joe the bread" \cite{li2022neural}. We believe that verbs are not strictly construction-specific, but according to our dataset and analysis, constructions tend to have slightly specific verbs. However, this does not mean they are limited to just those verbs and our result is limited to the dataset we used.

Previous studies have explored the processing of constructions in LLMs, but they often focused on specific types of constructions, resulting in limitations. For instance, Weissweiler's study concentrated solely on comparative correlative constructions \cite{weissweiler2023explaining}, Kyle Mahowald focused on Article + Adjective + Numeral + Noun (AANN) constructions \cite{mahowald2023discerning}, and Madabushi's research covered a broader range of constructions but did not specify which constructions were examined or how they relate to each other \cite{madabushi2020cxgbert}. Additionally, some studies used constructions with vastly different structures, making it less challenging for BERT to cluster them, and it is difficult to attribute this clustering to constructional differences \cite{weissweiler2023construction}\cite{veenboer2023using}\cite{xu2023enhancing}.

A recent study by Liu et al. stands out in this field, although its primary focus was on comparing verbs and constructions in sentence meaning rather than analyzing BERT's behavior \cite{liu2023quick}. Despite these contributions, there remains a gap in comprehensively understanding how LLMs process various types of constructions and how these constructions relate to each other. Additionally, Li et al.'s study used a dataset generated by a template, simplifying the clustering process. Consequently, the sentences often lack meaningful context, making it challenging to assess the behavior of natural language and the specificity of each token within specific constructions.

In our study, we decided to focus on argument structure constructions, as constructions in this family are similar, have most of the lexical units in common, and allow us to concentrate more on the constructional aspect of samples \cite{goldberg1995constructions}. These studies delve into the construction of BERT's hidden layer activity. Complementary to these works, we examine the attention heads in this model, as these heads are crucial components that could offer more detailed insights into the model's functionality. Attention mechanisms are inherently interpretable, as they indicate the extent to which a particular word influences the computation of the representation for the current word \cite{clark2019does}.

Research on attention heads has revealed that they follow limited patterns \cite{pande2021heads}, with much of the literature focused on defining the roles of these attention mechanisms \cite{pande2021heads}\cite{guan2020far}\cite{kovaleva2019revealing}. Given our focus on extracting features from attention mechanisms to understand how this system identifies constructions, our analysis will concentrate on the role of tokens. Tokens are easily traceable using multi-headed attention, making them an ideal focus for this investigation.

Our study also underscores the potential of transformer-based models to capture complex linguistic patterns in a manner that mirrors certain aspects of human language processing. The significant roles of the OBJ and VERB tokens in distinguishing ASCs suggest that these elements are critical in the syntactic and semantic parsing of sentences, a finding that could inform future research in both computational and cognitive neuroscience.

\subsection*{Possible Limitations and Future Directions}

While our analysis provides valuable insights, it is not without limitations. The reliance on synthetic data generated by GPT-4, while controlled, may not fully capture the complexities of natural language use. Future studies should consider using more diverse and naturally occurring datasets to validate these findings.

Additionally, while the FDR and GDV analyses offer quantitative measures of clustering and differentiation, further qualitative analysis is needed to understand the specific linguistic features that contribute to these patterns. Investigating the impact of different token types on ASC processing in more detail could reveal deeper insights into the underlying mechanisms.

A potential critique from a linguistic perspective might be that our study examines how one machine (BERT) processes language produced by another machine (GPT-4), which may not yield insights into natural language or how language is processed in the human brain. While this concern is valid, it is important to highlight that computational modeling is the first step towards understanding language processing in the brain. Using a controlled dataset generated by GPT-4 allows for clear differentiation between different Argument Structure Constructions (ASCs) and removes confounding variables present in natural language, enabling a more focused study of BERT's processing capabilities.

Furthermore, GPT-4 is trained on one of the largest and most diverse language corpora ever assembled, making its generated datasets equally valid as language corpora. This extensive training allows GPT-4 to produce language that mirrors the statistical properties of natural language, capturing a wide range of linguistic phenomena. As such, analyzing how BERT processes GPT-4-generated language can still provide meaningful insights into the fundamental principles of language processing.

Furthermore, the results obtained from our study align with established linguistic theories and findings from studies using natural language, suggesting that the underlying principles captured by these models are relevant. Additionally, future work will involve validating these findings with naturally occurring datasets and comparing them with neuroimaging data to better understand the parallels between computational models and human brain processing. Thus, while recognizing the limitations, our study provides a foundational step toward bridging the gap between artificial and natural language processing, contributing valuable insights to both computational linguistics and cognitive neuroscience.

\subsection*{Conclusion}

In conclusion, BERT effectively captures both the specific and general aspects of grammatical constructions, with its layers progressively integrating lexical, syntactic, and semantic information. This study demonstrates BERT's nuanced understanding of linguistic structures, albeit with certain challenges in differentiating closely related constructions like transitive and resultative sentences.

Our study highlights the sophisticated capabilities of the BERT language model in representing and differentiating between various Argument Structure Constructions. The dynamic and layered nature of BERT's processing, as revealed through clustering, probing, and attention weight analyses, underscores the model's potential to mirror human linguistic processing.

Future research aimed at comparing these computational representations with neuroimaging data will be pivotal in advancing our understanding of the computational and neural mechanisms underlying language comprehension. In particular, comparing our computational findings with neuroimaging data during continuous speech perception will be crucial in bridging the gap between computational models and the neural mechanisms of language understanding. Such comparisons could validate whether the patterns observed in BERT align with how the human brain processes different ASCs, offering a more comprehensive view of language processing.


\section*{Author contributions}
All authors discussed the results and approved the final version of the manuscript.

\section*{Acknowledgements}
This work was funded by the Deutsche Forschungsgemeinschaft (DFG, German Research Foundation): KR\,5148/3-1 (project number 510395418), KR\,5148/5-1 (project number 542747151), and GRK\,2839 (project number 468527017) to PK, and grant SCHI\,1482/3-1 (project number 451810794) to AS.



\end{document}